\documentclass{article}

\PassOptionsToPackage{numbers, compress}{natbib}
\usepackage[final]{nips_2016}

\usepackage[utf8]{inputenc} 
\usepackage[T1]{fontenc}    
\usepackage[hidelinks]{hyperref}       
\usepackage{url}            
\usepackage{booktabs}       
\usepackage{amsfonts}       
\usepackage{nicefrac}       
\usepackage{microtype}      

\usepackage[caption=false]{subfig}
\usepackage[export]{adjustbox}
\usepackage[]{nth}
\usepackage[flushleft]{threeparttable} 
\usepackage{tabularx} 
\usepackage{color}
\usepackage[cmex10]{amsmath}
\usepackage{adjustbox}

\newcommand{\minisection}[1]{\vspace{0.04in} \noindent {\bf #1}\ \ }

\newcommand\blfootnote[1]{%
  \begingroup
  \renewcommand\thefootnote{}\footnote{#1}%
  \addtocounter{footnote}{-1}%
  \endgroup
}

\title{Invertible Conditional GANs for image editing}

\author{
    Guim Perarnau, Joost van de Weijer, Bogdan Raducanu \\
  Computer Vision Center\\
  Barcelona, Spain \\
  \texttt{guimperarnau@gmail.com} \\
  \texttt{\{joost,bogdan\}@cvc.uab.es}
  \And
   Jose M. \'Alvarez \\
   Data61 @ CSIRO\\
   Canberra, Australia \\
  \texttt{jose.alvarez@nicta.com.au} \\
}

\begin{document}
\blfootnote{Code available at \url{https://github.com/Guim3/IcGAN}}
\maketitle

\begin{abstract}
Generative Adversarial Networks (GANs) have recently demonstrated to successfully approximate complex data distributions. A relevant extension of this model is conditional GANs (cGANs), where the introduction of external information allows to determine specific representations of the generated images. In this work, we evaluate encoders to inverse the mapping of a cGAN, i.e., mapping a real image into a latent space and a conditional representation. This allows, for example, to reconstruct and modify real images of faces conditioning on arbitrary attributes. Additionally, we evaluate the design of cGANs. The combination of an encoder with a cGAN, which we call Invertible cGAN (IcGAN), enables to re-generate real images with deterministic complex modifications.
\end{abstract}

\section{Introduction} \label{sec:intro}
Image editing can be performed at different levels of complexity and abstraction. Common operations consist in simply applying a filter to an image to, for example, augment the contrast or convert to grayscale. These, however, are low-complex operations that do not necessarily require to comprehend the scene or object that the image is representing. On the other hand, if one would want to modify the attributes of a face (e.g. add a smile, change the hair color or even the gender), this is a more complex and challenging modification to perform. In this case, in order to obtain realistic results, a skilled human with an image edition software would often be required.

A solution to automatically perform these non-trivial operations relies on generative models. Natural image generation has been a strong research topic for many years, but it has not been until 2015 that promising results have been achieved with deep learning techniques combined with generative modeling \citep{Gregor2015,Radford2015}. Generative Adversarial Networks (GANs) \cite{Goodfellow2014} is one of the state-of-the-art approaches for image generation. GANs are especially interesting as they are directly optimized towards generating the most plausible and realistic data, as opposed to other models (e.g. Variational Autoencoders \cite{Kingma2013}), which focus on an image reconstruction loss. Additionally, GANs are able to explicitly control generated images features with a conditional extension, conditional GANs (cGANs). However, the GAN framework lacks an inference mechanism, i.e., finding the latent representation of an input image, which is a necessary step for being able to reconstruct and modify real images.

\begin{figure}[t]
        \centering
        \includegraphics[width=0.85\linewidth]{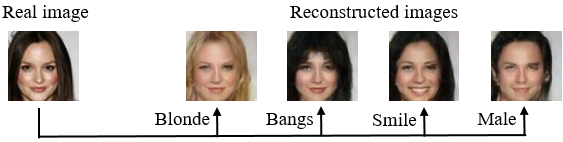}
        \caption{Example of how the IcGAN reconstructs and applies complex variations on a real image.}
        \label{fig:icgan_example}
\end{figure}

In order to overcome this limitation, in this paper we introduce Invertible Conditional GANs (IcGANs) for complex image editing as the union of an encoder used jointly with a cGAN. This model allows to map real images into a high-feature space (encoder) and perform meaningful modifications on them (cGAN). As a result, we can explicitly control the attributes of a real image (Figure \ref{fig:icgan_example}), which could be potentially useful in several applications, be it creative processes, data augmentation or face profiling.

\begin{minipage}{\linewidth} 
The summary of contributions of our work is the following:
\begin{itemize}
\item Proposing IcGANs, composed of two crucial parts: an encoder and a cGAN. We apply this model to MNIST \citep{LeCun1998} and CelebA \citep{celeba} datasets, which allows performing meaningful and realistic editing operations on them by arbitrarily changing the conditional information $y$.   
\item Introducing an encoder in the conditional GAN framework to compress a real image $x$ into a latent representation $z$ and conditional vector $y$. We consider several designs and training procedures to leverage the performance obtained from available conditional information.
\item Evaluating and refining cGANs through conditional position and conditional sampling to enhance the quality of generated images.
\end{itemize}
\end{minipage}

\section{Related work} \label{sec:state_of_art}
There are different approaches for generative models. Among them, there are two promising ones that are recently pushing the state-of-the-art with highly plausible generated images.

The first one is Variational Autoencoders (VAE) \citep{Gregor2015,Kingma2013,Rezende2014,Kingma2014}, which impose a prior representation space $z$ (e.g. normal distribution) in order to regularize and constrain the model to sample from it.
However, VAEs main limitation is the pixel-wise reconstruction error used as a loss function, which causes the output images to look blurry.
The second approach is Generative Adversarial Nets (GANs). Originally proposed by Goodfellow \textit{et al.} \citep{Goodfellow2014}, GANs have been improved with a deeper architecture (DCGAN) by Radford \textit{et al.} \citep{Radford2015}. The latest advances introduced several techniques that improve the overall performance for training GANs \citep{Salimans2016} and an unsupervised approach to disentangle feature representations \citep{Chen2016}.
Additionally, the most advanced and recent work on cGANs trains a model to generate realistic images from text descriptions and landmarks \cite{Reed2016_2}.


Our work is considered in the content of the GAN framework. The baseline will be the work of Radford's \textit{et al.} (DCGANs) \citep{Radford2015}, which we will add a conditional extension. The difference of our approach to prior work is that we also propose an encoder (Invertible cGAN) with which we can, given an input image $x$, to obtain its representation as a latent variable $z$ and a conditional vector $y$. Then, we can modify $z$ and $y$ to re-generate the original image with complex variations. 
Dumoulin \textit{et al.} \citep{Dumoulin2016} and Donahue \textit{et al.} \citep{Donahue2016} also proposed an encoder in GANs, but in a non-conditional and jointly trained setting.
Additionally, Makhzani \textit{et al.} \citep{Makhzani2015} and Larsen \textit{et al.} \citep{Larsen2015} proposed a similar idea to this paper by combining a VAE and a GAN with promising results.

Reed \textit{et al.} \citep{Reed2016} implemented an encoder in a similar fashion to our approach. This paper builds alongside their work in a complementary manner. In our case, we analyze more deeply the encoder by including conditional information encoding and testing different architectures and training approaches. Also, we evaluate unexplored design decisions for building a cGAN.

\section{Background: Generative Adversarial Networks} \label{sec:GANs}
A GAN is composed of two neural networks, a generator $G$ and a discriminator $D$. Both networks are iteratively trained competing against each other in a minimax game. The generator aims to approximate the underlying unknown data distribution $p_{data}$ to fool the discriminator, whilst the discriminator is focused on being able to tell which samples are real or generated. On convergence, we want $p_{data}=p_g$, where $p_g$ is the generator distribution. 


More formally, considering the function $v(\theta_g,\theta_d)$, where $\theta_g$ and $\theta_d$ are the parameters of the generator $G$ and discriminator $D$ respectively, we can formulate GAN training as optimizing

\begin{equation}
\operatornamewithlimits{min}\limits_{g} \operatornamewithlimits{max}\limits_{d} v(\theta_g,\theta_d) = \mathbb{E}_{x\sim p_{data}} [\log D(x)] + \mathbb{E}_{z\sim p_z}[ \log(1-D(G(z)))],
\label{eq:GAN}
\end{equation}
where $z$ is a vector noise sampled from a known simple distribution $p_z$ (e.g. normal).
 
GAN framework can be extended with conditional GANs (cGANs) \citep{Mirza2014}. They are quite similar to vanilla (non-conditional) GANs, the only difference is that, in this case, we have extra information $y$ (e.g. class labels, attribute information) for a given real sample $x$. Conditional information strictly depends on real samples, but we can model a density model $p_y$ in order to sample generated labels $y'$ for generated data $x'$. Then, Equation \ref{eq:GAN} can be reformulated for the cGAN extension as

\begin{equation}
\operatornamewithlimits{min}\limits_{g} \operatornamewithlimits{max}\limits_{d} v(\theta_g,\theta_d) = \mathbb{E}_{x,y\sim p_{data}} [\log D(x,y)] + \mathbb{E}_{z\sim p_z, y'\sim p_y}[ \log(1-D(G(z,y'),y'))].
\label{eq:cGAN}
\end{equation} 


Once a cGAN is trained, it allows us to generate samples using two level of variations: constrained and unconstrained. Constrained variations are modeled with $y$ as it directly correlates with features of the data that are explicitly correlated with $y$ and the data itself. Then, all the other variations of the data not modeled by $y$ (unconstrained variations) are encoded in $z$.

\section{Invertible Conditional GANs} \label{sec:IcGANs}
We introduce Invertible Conditional GANs (IcGANs), which are composed of a cGAN and an encoder. Even though encoders have recently been introduced into the GAN framework \cite{Dumoulin2016,Donahue2016,Reed2016}, we are the first ones to include and leverage the conditional information $y$ into the design of the encoding process. In section \ref{icgan:enc} we explain how and why an encoder is included in the GAN framework for a conditional setting. In section \ref{icgan:cgan}, we introduce our approach to refine cGANs on two aspects: conditional position and conditional sampling. The model architecture is described in section \ref{icgan:arch}.

\subsection{Encoder} \label{icgan:enc}
\begin{figure}[t]
        \centering
        \includegraphics[width=0.92\linewidth]{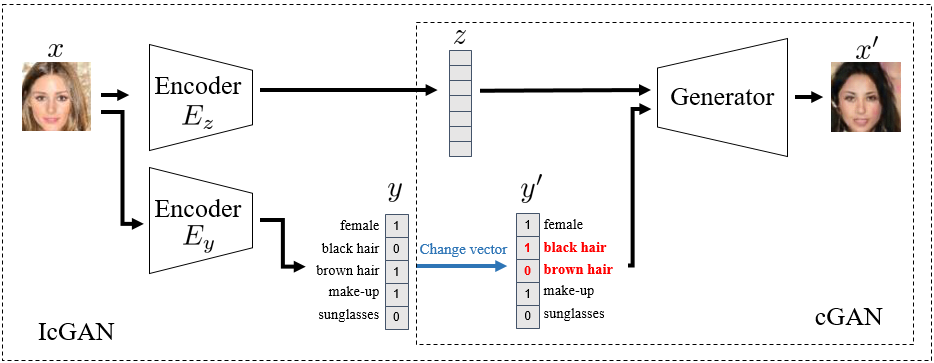}
        \caption{Scheme of a trained IcGAN, composed of an encoder (IND approach) and a cGAN generator. We encode a real image $x$ into a latent representation $z$ and attribute information $y$, and then apply variations on it to generate a new modified image $x'$.}
        \label{fig:icgan_overview}
\end{figure}

A generator $x' = G(z,y')$ from a GAN framework does not have the capability to map a real image $x$ to its latent representation $z$. To overcome this problem, we can train an encoder/inference network $E$ that approximately inverses this mapping $(z,y) = E(x)$. This inversion would allow us to have a latent representation $z$ from a real image $x$ and, then, we would be able to explore the latent space by interpolating or adding variations on it, which would result in variations on the generated image $x'$. If combined with a cGAN, once the latent representation $z$ has been obtained, explicitly controlled variations can be added to an input image via conditional information $y$ (e.g. generate a certain digit in MNIST or specify face attributes on a face dataset). We call this combination Invertible cGAN, as now the mapping can be inverted: $(z,y) = E(x)$ and $x' = G(z,y)$, where $x$ is an input image and $x'$ its reconstruction. See Figure \ref{fig:icgan_overview} for an example on how a trained IcGAN is used.

Our approach consists of training an encoder $E$ once the cGAN has been trained, as similarly considered by Reed \textit{et al} \cite{Reed2016}. In our case, however, the encoder $E$ is composed of two sub-encoders: $E_z$, which encodes an image to $z$, and $E_y$, which encodes an image to $y$. To train $E_z$ we use the generator to create a dataset of generated images $x'$ and their latent vectors $z$, and then minimize a squared reconstruction loss $\mathcal{L}_{ez}$ (Eq. \ref{eq:Ez}). For $E_y$, we initially used generated images $x'$ and their conditional information $y'$ for training. However, we found that generated images tend to be noisier than real ones and, in this specific case, we could improve $E_y$ by directly training with real images and labels from the dataset $p_{data}$ (Eq. \ref{eq:Ey}).
\begin{flalign}
&\mathcal{L}_{ez}=\mathbb{E}_{z\sim p_z, y'\sim p_y}\|z-E_z(G(z,y'))\|^2_2 \label{eq:Ez} \\ 
&\mathcal{L}_{ey}=\mathbb{E}_{x,y\sim p_{data}}\|y-E_y(x)\|^2_2 \label{eq:Ey}
\end{flalign} 

Although $E_z$ and $E_y$ might seem completely independent, we can adopt different strategies to make them interact and leverage the conditional information (for an evaluation of them, see section \ref{exp:enc_eval}): 
\begin{itemize}
\item SNG: One single encoder with shared layers and two outputs. That is, $E_z$ and $E_y$ are embedded in a single encoder.
\item IND: Two independent encoders. $E_z$ and $E_y$ are trained separately.
\item IND-COND: Two encoders, where $E_z$ is conditioned on the output of encoder $E_y$.
\end{itemize}

Recently, Dumoulin \textit{et al.} \citep{Dumoulin2016} and Donahue \textit{et al.} \citep{Donahue2016} proposed different approaches on how to train an encoder in the GAN framework. One of the most interesting approaches consists in jointly training the encoder with both the discriminator and the generator.
Although this approach is promising, our work has been completely independent of these articles and  focuses on another direction, since we consider the encoder in a conditional setting. Consequently, we implemented our aforementioned approach which performs nearly equally \citep{Donahue2016} to their strategy.

\subsection{Conditional GAN} \label{icgan:cgan}
We consider two main design decisions concerning cGANs. The first one is to find the optimal conditional position $y$ on the generator and discriminator, which, to our knowledge, has not been previously addressed. Secondly, we discuss the best approach to sample conditional information for the generator.

\minisection{Conditional position}
In the cGAN, the conditional information vector $y$ needs to be introduced in both the generator and the discriminator. In the generator, $y \sim p_{data}$ and $z \sim p_z$ (where $p_z=\mathcal{N}(0,1))$ are always concatenated in the filter dimension at the input level \citep{Reed2016,Mirza2014,Gauthier2014}. As for the discriminator, different authors insert $y$ in different parts of the model \citep{Reed2016,Mirza2014,Gauthier2014}. We expect that the earlier $y$ is positioned in the model the better since the model is allowed to have more learning interactions with $y$. Experiments regarding the optimal $y$ position will be detailed in section \ref{subsec:cGAN_exp}.

\minisection{Conditional sampling} There are two types of conditional information, $y$ and $y'$. The first one is trivially sampled from $(x,y) \sim p_{data}$ and is used for training the discriminator $D(x,y)$ with a real image $x$ and its associated label $y$. The second one is sampled from $y' \sim p_y$ and serves as input to the generator $G(z,y')$ along with a latent vector $z \sim p_z$ to generate an image $x'$, and it can be sampled using different approaches:
\begin{itemize}
\item Kernel density estimation: also known as Parzen window estimation, it consists in randomly sampling from a kernel (e.g. Gaussian kernel with a cross-validated $\sigma$). 
\item Direct interpolation: interpolate between label vectors $y$ from the training set \citep{Reed2016}. The reasoning behind this approach is that interpolations can belong to the label distribution $p_y$.
\item Sampling from the training set $y' \sim p_y$, $p_y=p_{data}$: Use directly the real labels $y$ from the training set $p_{data}$. As Gauthier \citep{Gauthier2014} pointed out, unlike the previous two approaches, this method could overfit the model by using the conditional information to reproduce the images of the training set. However, this is only likely to occur if the conditional information is, to some extent, unique for each image. In the case where the attributes of an image are binary, one attribute vector $y$ could describe a varied and large enough subset of images, preventing the model from overfitting given $y$.
\end{itemize}

Kernel density estimation and direct interpolation are, at the end, two different ways to interpolate on $p_y$. Nevertheless, interpolation is mostly suitable when the attribute information $y$ is composed of real vectors $y \in \mathbb{R}^n$, not binary ones. It is not the case of the binary conditional information of the datasets used in this paper (see section \ref{subsec:data} for dataset information). Directly interpolating binary vectors would not create plausible conditional information, as an interpolated vector $y\in \mathbb{R}^n$ would not belong to $p_y \in \{0,1\}^n$ nor $p_{data}\in\{0,1\}^n$. Using a kernel density estimation would not make sense either, as all the binary labels would fall in the corners of a hypercube. Therefore, we will directly sample $y$ from $p_{data}$.

\subsection{Model architecture} \label{icgan:arch}
\begin{figure}[t]
        \centering
        \subfloat[]{
                \includegraphics[width=0.50\linewidth]{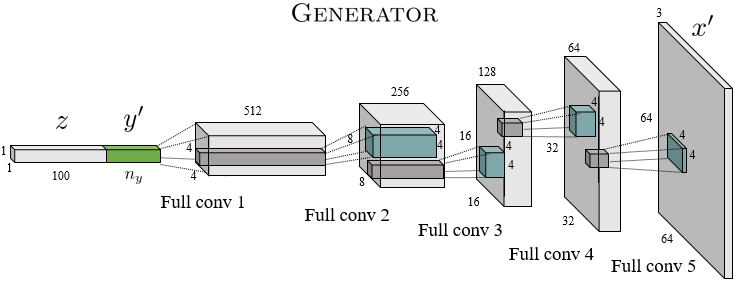}
        }
        \subfloat[]{
                \includegraphics[width=0.475\linewidth]{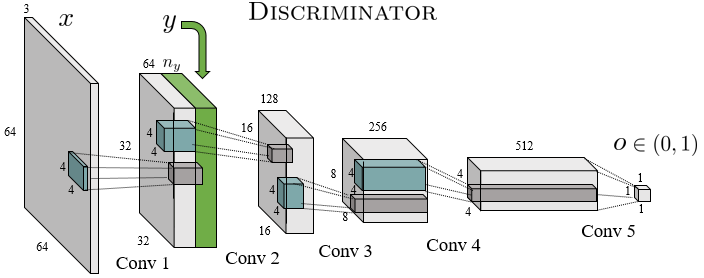}

        }
        \caption{Architecture of the generator (a) and discriminator (b) of our cGAN model. The generator $G$ takes as input both $z$ and $y$. In the discriminator, $y$ is concatenated in the first convolutional layer.}\label{fig:cGAN_arch}
\end{figure}

\minisection{Conditional GAN}
The work of this paper is based on the Torch implementation of the DCGAN\footnote{Torch code for DCGAN model available at \url{https://github.com/soumith/dcgan.torch}} \cite{Radford2015}. We use the recommended configuration for the DCGAN, which trains with the Adam optimizer \citep{adam2014} ($\beta_1=0.5,\beta_2=0.999,\epsilon=10^{-8}$) with a learning rate of $0.0002$ and a mini-batch size of $64$ (samples drawn independently at each update step) during $25$ epochs. The output image size used as a baseline is $64\times64$. Also, we train the cGAN with the matching-aware discriminator method from Reed \textit{et al.} \cite{Reed2016}.
In Figure \ref{fig:cGAN_arch} we show an overview architecture of both generator and discriminator for the cGAN. For a more detailed description of the model see Table \ref{tab:arch}.

\begin{table}[ht]
\centering
\caption{Detailed generator and discriminator architecture}
\label{tab:arch}
\begin{adjustbox}{width=1.0\textwidth}
\begin{tabular}{llllll|llllll}
\multicolumn{6}{c|}{\textbf{Generator}}                                                                       & \multicolumn{6}{c}{\textbf{Discriminator}}                                                                                                                                             \\
\textbf{Operation} & \textbf{Kernel} & \textbf{Stride} & \textbf{Filters} & \textbf{BN} & \textbf{Activation} & \textbf{Operation}              & \textbf{Kernel}             & \textbf{Stride}             & \textbf{Filters}             & \textbf{BN}             & \textbf{Activation}             \\
\multicolumn{6}{l|}{Concatenation \qquad\textit{Concatenate $z$ and $y'$ on \nth{1} dimension}}                            & Convolution                     & $4\times4$                  & $2\times2$                  & 64                           & No                      & Leaky ReLU                      \\
Full convolution   & $4\times4$      & $2\times2$      & 512              & Yes         & ReLU                & \multicolumn{6}{l}{Concatenation  \thinspace\thinspace \textit{Replicate $y$ and concatenate to \nth{1} conv. layer}} \\
Full convolution   & $4\times4$      & $2\times2$      & 256              & Yes         & ReLU                & Convolution                     & $4\times4$                  & $2\times2$                  & 128                          & Yes                     & Leaky ReLU                      \\
Full convolution   & $4\times4$      & $2\times2$      & 128              & Yes         & ReLU                & Convolution                     & $4\times4$                  & $2\times2$                  & 256                          & Yes                     & Leaky ReLU                      \\
Full convolution   & $4\times4$      & $2\times2$      & 64               & Yes         & ReLU                & Convolution                     & $4\times4$                  & $2\times2$                  & 512                          & Yes                     & Leaky ReLU                      \\
Full convolution   & $4\times4$      & $2\times2$      & 3                & No          & Tanh                & Convolution                     & $4\times4$                  & $1\times1$                  & 1                            & No                      & Sigmoid                        
\end{tabular}
\end{adjustbox}
\end{table}
\minisection{Encoder}
For simplicity, we show the architecture of the IND encoders (Table \ref{tab:enc_arch}), as they are the ones that give the best performance. Batch Normalization and non-linear activation functions are removed from the last layer to guarantee that the output distribution is similar to $p_z = \mathcal{N}(0,1)$. Additionally, after trying different configurations, we have replaced the last two convolutional layers with two fully connected layers at the end of the encoder, which yields a lower error. The training configuration (Adam optimizer, batch size, etc) is the same as the one used for the cGAN model.  

\begin{table}[ht]
\centering
\caption{Encoder IND architecture. Last two layers have different sizes depending on the encoder ($z$ for $E_z$ or $y$ for $E_y$). $n_y$ represents the size of $y$.} 
\label{tab:enc_arch}
\begin{adjustbox}{width=0.6\textwidth}
\begin{tabular}{llllll}
\textbf{Operation} & \textbf{Kernel} & \textbf{Stride} & \textbf{Filters} & \textbf{BN} & \textbf{Activation} \\
Convolution        & $5\times5$      & $2\times2$      & 32               & Yes         & ReLU                \\
Convolution        & $5\times5$      & $2\times2$      & 64              & Yes         & ReLU                \\
Convolution        & $5\times5$      & $2\times2$      & 128              & Yes         & ReLU                \\
Convolution        & $5\times5$      & $2\times2$      & 256              & Yes         & ReLU                \\
Fully connected    & -               & -               & $z$: 4096, $y$: 512           & Yes         & ReLU                \\
Fully connected    & -               & -               & $z$: 100, $y$: $\mathit{n_y}$                & No          & None      \\
\end{tabular}
\end{adjustbox}
\end{table}

\section{Experiments} \label{sec:experiments}
\subsection{Datasets} \label{subsec:data}
We use two image datasets of different complexity and variation, MNIST \citep{LeCun1998} and CelebFaces Attributes (CelebA) \citep{celeba}. MNIST is a digit dataset of grayscale images composed of 60,000 training images and 10,000 test images. Each sample is a $28\times28$ centered image labeled with the class of the digit (0 to 9). CelebA is a dataset composed of 202,599 face colored images and 40 attribute binary vectors. We use the aligned and cropped version and scale the images down to $64\times64$. We also use the official train and test partitions, 182K for training and 20K for testing. Of the original 40 attributes, we filter those that do not have a clear visual impact on the generated images, which leaves a total of 18 attributes. 
We will evaluate the quality of generated samples of both datasets. However, a quantitative evaluation will be performed on CelebA only, as it is considerably more complex than MNIST. 

\subsection{Evaluating the conditional GAN} \label{subsec:cGAN_exp}
The goals of this experiment are two. First, we evaluate the general performance of the cGAN with an attribute predictor network (Anet) on CelebA dataset. Second, we test the impact of adding $y$ in different layers of the cGAN (section \ref{icgan:cgan}, conditional position). 

We use an Anet\footnote{The architecture of the Anet is the same as $E_y$ from Table \ref{tab:enc_arch}.} as a way to make a quantitative evaluation in a similar manner as Salimans \textit{et al.} Inception model \cite{Salimans2016}, as the output given by this Anet (i.e., which attributes are detected on a generated sample) is a good indicator of the generator ability to model them. In other words, if the predicted Anet attributes $y'$ are closer to the original attributes $y$ used to generate an image $x'$, we expect that the generator has successfully learned the capability to generate new images considering the semantic meaning of the attributes. Therefore, we use the generator $G$ to create images $x'$ conditioned on attribute vectors $y \sim p_{data}$ (i.e. $x' = G(z,y)$), and make the Anet predict them. Using the Anet output, we build a confusion matrix for each attribute and compute the mean accuracy and F1-Score to test the model and its inserted optimal position of $y$ in both generator and discriminator.

\begin{table}[ht]
\centering
\caption{Quantitative cGAN evaluation depending on $y$ inserted position. The first row shows the results obtained with real CelebA images as an indication that Anet predictions are subject to error.}
\label{tab:cGAN_quantEval}
\begin{adjustbox}{width=0.80\textwidth}
\begin{tabular}{lllll}
                        & \multicolumn{2}{c}{\textbf{Discriminator}}                           & \multicolumn{2}{c}{\textbf{Generator}}          \\
\textbf{Model}          & \textbf{Mean accuracy} & \multicolumn{1}{l|}{\textbf{Mean F1-Score}} & \textbf{Mean accuracy} & \textbf{Mean F1-Score} \\
CelebA test set         & 92.78\%                & \multicolumn{1}{l|}{71.47\%}                & 92.78\%                & 71.47\%                \\
$y$ inserted on input            & 85.74\%                & \multicolumn{1}{l|}{49.63\%}                & \textbf{89.83\%}       & \textbf{59.69\%}       \\
$y$ inserted on layer 1 & \textbf{86.01\%}       & \multicolumn{1}{l|}{\textbf{52.42\%}}       & 87.16\%                & 52.40\%                \\
$y$ inserted on layer 2 & 84.90\%                & \multicolumn{1}{l|}{50.00\%}                & 82.49\%                & 52.36\%                \\
$y$ inserted on layer 3 & 85.96\%                & \multicolumn{1}{l|}{52.38\%}                & 82.49\%                & 38.01\%                \\
$y$ inserted on layer 4 & 77.61\%                & \multicolumn{1}{l|}{19.49\%}                & 73.90\%                & 4.03\%                
\end{tabular}
\end{adjustbox}
\end{table}

In Table \ref{tab:cGAN_quantEval} we can see how cGANs have successfully learned to generate the visual representations of the conditional attributes with an overall accuracy of $\sim86$\%. The best accuracy is achieved by inserting $y$ in the first convolutional layer of the discriminator and at the input level for the generator. Thus, we are going to use this configuration for the IcGAN. Both accuracy and F1-Score are similar as long as $y$ is not inserted in the last convolutional layers, in which case the performance considerably drops, especially in the generator. Then, these results reinforce our initial intuition of $y$ being added at an early stage of the model to allow learning interactions with it.

\subsection{Evaluating the encoder} \label{exp:enc_eval}
\begin{figure}[t]
        \centering
        \subfloat[]{
                \includegraphics[width=0.3255\linewidth]{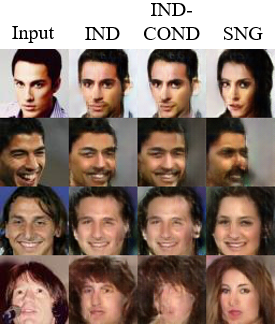}
        }
        \qquad
        \subfloat[]{
                \includegraphics[width=0.525\linewidth]{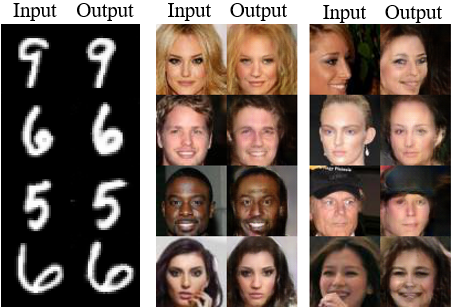}
}
        \caption{(a) Comparison of different encoder configurations, where IND yields the most faithful reconstructions. (b) Reconstructed samples from MNIST and CelebA using IND configuration.}
        \label{fig:encoder_samples}
\end{figure}

In this experiment, we prioritize the visual quality of reconstructed samples as an evaluation criterion. Among the different encoder configurations of section \ref{icgan:enc}, IND and IND-COND yield a similar qualitative performance, being IND slightly superior. A comparison of these different configurations is shown in Figure \ref{fig:encoder_samples}a and in Figure \ref{fig:encoder_samples}b we focus on IND reconstructed samples. 
On another level, the fact that the generator is able, via an encoder, to reconstruct unseen images from the test set shows that the cGAN is generalizing and suggests that it does not suffer from overfitting, i.e., it is not just memorizing and reproducing training samples. 

Additionally, we compare the different encoder configurations in a quantitative manner by using the minimal squared reconstruction loss $\mathcal{L}_e$ as a criterion. Each encoder is trained minimizing $\mathcal{L}_{e}$ with respect to latent representations $z$ ($\mathcal{L}_{ez}$) or conditional information $y$ ($\mathcal{L}_{ey}$). Then, we quantitatively evaluate different model architectures using $\mathcal{L}_e$ as a metric on a test set of 150K CelebA generated images. We find that the encoder that yields the lowest $\mathcal{L}_e$ is also IND (0.429), followed closely by IND-CND (0.432), and being SNG the worst case (0.500). 

Furthermore, we can see an interesting property of minimizing a loss based on the latent space instead of a pixel-wise image reconstruction: reconstructed images tend to accurately keep high-level features of an input image (e.g. how a face generally looks) in detriment to more local details such as the exact position of the hair, eyes or face. Consequently, a latent space based encoder is invariant to these local details, making it an interesting approach for encoding purposes. For example, notice how the reconstructions in the last row of CelebA samples in Figure \ref{fig:encoder_samples}b fill the occluded part of the face by a hand. Another advantage with respect to element-wise encoders such as VAE is that GAN based reconstructions do not look blurry.

\subsection{Evaluating the IcGAN}

\begin{figure}[t]
		\centering
        \subfloat[]{
                \includegraphics[width=0.95\linewidth]{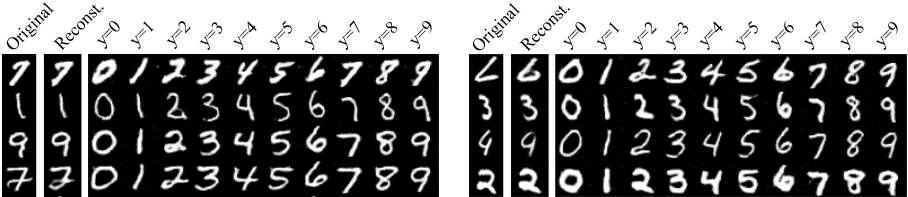}
        }
        \newline
        \subfloat[]{
                \includegraphics[width=0.95\linewidth]{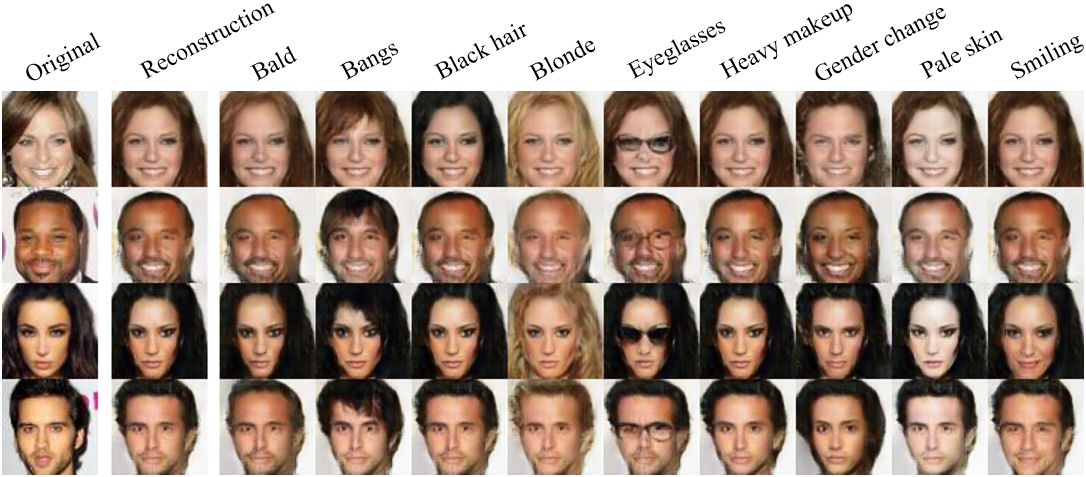}
	    }
        \caption{The result of applying an IcGAN to a set of real images from MNIST (a) and CelebA (b). A real image is encoded into a latent representation $z$ and conditional information $y$, and then decoded into a new image. We fix $z$ for every row and modify $y$ for each column to obtain variations.} \label{fig:icgan_analogies}
\end{figure}

In order to test that the model is able to correctly encode and re-generate a real image by preserving its main attributes, we take real samples from MNIST and CelebA test sets and reconstruct them with modifications on the conditional information $y$. The result of this procedure is shown in Figure \ref{fig:icgan_analogies}, where we show a subset of 9 of the 18 for CelebA attributes for image clarity. We can see that, in MNIST, we are able to get the hand-written style of real unseen digits and replicate these style on all the other digits. On the other hand, in CelebA we can see how reconstructed faces generally match the specified attribute. Additionally, we noticed that faces with uncommon conditions (e.g., looking away from the camera, face not centered) were the most likely to be noisy. Furthermore, attributes such as \textit{mustache} often fail to be generated especially on women samples, which might indicate that the generator is limited to some unusual attribute combinations. 

\begin{figure}[t]
        \centering
        \subfloat[]{
                \includegraphics[width=0.555\linewidth]{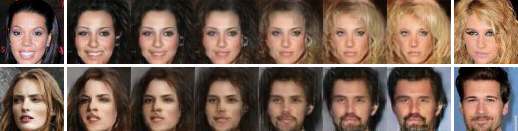}
        }
        \subfloat[]{
                \includegraphics[width=0.44\linewidth]{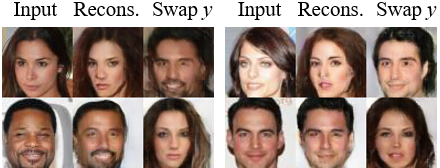}
	    }
        \caption{Different ways of exploring the latent space. In (a) we take two real images and linearly interpolate both $z$ and $y$ to obtain a gradual transformation from one face to another. In (b) we take two real images, reconstruct them and swap the attribute information $y$ between them.} \label{fig:icgan_manipulate_Z}
\end{figure}

\minisection{Manipulating the latent space} \label{exp:icgan:latent}
The latent feature representation $z$ and conditional information $y$ learned by the generator can be further explored beyond encoding real images or randomly sampling $z$. In order to do so, we linearly interpolate both $z$ and $y$ with pairs of reconstructed images from the CelebA test set (Figure \ref{fig:icgan_manipulate_Z}a). All the interpolated faces are plausible and the transition between faces is smooth, demonstrating that the IcGAN learned manifold is also consistent between interpolations. Then, this is also a good indicator that the model is generalizing the face representation properly, as it is not directly memorizing training samples.

In addition, we perform in Figure \ref{fig:icgan_manipulate_Z}b an attribute transfer between pairs of faces. We infer the latent representation $z$ and attribute information $y$ of two real faces from the test set, swap $y$ between those faces and re-generate them. As we previously noticed, the results suggest that $z$ encodes pose, illumination and background information, while $y$ tends to represent unique features of the face.

\section{Conclusions} \label{sec:conclusions}

We introduce an encoder in a conditional setting within the GAN framework, a model which we call Invertible Conditional GANs (IcGANs). It solves the problem of GANs lacking the ability to infer real samples to a latent representation $z$, while also allowing to explicitly control complex attributes of generated samples with conditional information $y$. We also refine the performance of cGANS by testing the optimal position in which the conditional information $y$ is inserted in the model. We have found that for the generator, $y$ should be added at the input level, whereas the discriminator works best when $y$ is at the first layer. Additionally, we evaluate several ways to training an encoder. Training two independent encoders -- one for encoding $z$ and another for encoding $y$ -- has proven to be the best option in our experiments. The results obtained with a complex face dataset, CelebA, are satisfactory and promising.


\minisection{Acknowledgments}
This work is funded by the Projects TIN2013-41751-P of the Spanish Ministry of Science and the CHIST ERA project PCIN-2015-226.

\newpage
\bibliographystyle{IEEEtranN}
{\small
\bibliography{references.bib}}

\end{document}